\documentclass[11pt]{article}

\usepackage[preprint]{acl}

\usepackage{times}
\usepackage{latexsym}

\usepackage{microtype}
\usepackage{inconsolata}

\usepackage{graphicx}

\usepackage{amsmath}
\usepackage{amssymb}
\usepackage{amsfonts}
\usepackage{amsthm}
\usepackage{bbm}

\usepackage{booktabs}
\usepackage{multirow}
\usepackage{xcolor}
\usepackage{colortbl}
\usepackage{tcolorbox}

\usepackage{algorithm}
\usepackage{algorithmic}

\usepackage[T1]{fontenc}

\usepackage[utf8]{inputenc}

\usepackage{microtype}
\usepackage{inconsolata}
\usepackage{graphicx}

\usepackage[misc]{ifsym}
\usepackage{fancyhdr}

\makeatletter
\fancypagestyle{firstpage}{
  \fancyhf{}
  \renewcommand{\headrulewidth}{0.4pt}
  \IfFileExists{hunyuanlogo.png}{
    \lhead{\includegraphics[height=18pt]{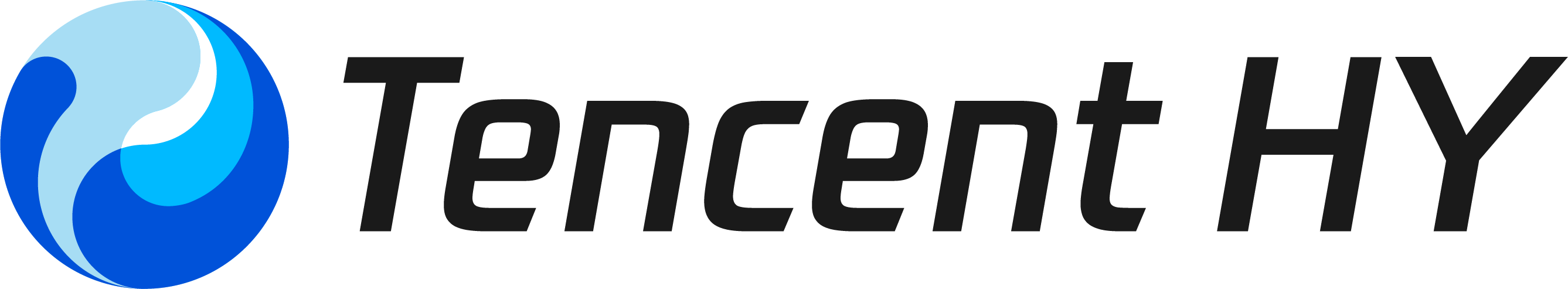}}
  }{}
  \rhead{\textcolor[HTML]{4D4D4D}{\small \today}}
  \renewcommand{\headrule}{{\color[HTML]{4D4D4D}\hrule\@height\headrulewidth\@width\headwidth\vskip-\headrulewidth}}
  \fancyfoot[C]{\thepage}
}
\makeatother

\title{ADWIN: Adaptive Windows for Horizon-Aware On-Policy Distillation}

\author{
    \textbf{Kun Liang}$^{1,2,*}$ \quad
    \textbf{Chenming Tang}$^{1,2,*}$ \quad
    \textbf{Clive Bai}$^{3}$ \\
    \textbf{Weijie Liu}$^{3}$  \quad
    \textbf{Saiyong Yang}$^{3,\dagger}$ \quad
    \textbf{Yunfang Wu}$^{1,2,\dagger}$ \\
    \noalign{\vskip 5pt}
    $^1$School of Computer Science, Peking University \\
    $^2$National Key Laboratory for Multimedia Information Processing, Peking University \\
    $^3$LLM Department, Tencent \\
    \noalign{\vskip 5pt}
    \Letter~{\fontsize{10.5}{12}\selectfont\texttt{kliang25@stu.pku.edu.cn}} \hspace{0.5em}
    \Letter~{\fontsize{10.5}{12}\selectfont\texttt{wuyf@pku.edu.cn}}
}

\begin{document}
\maketitle
\thispagestyle{firstpage}

\renewcommand*{\thefootnote}{\fnsymbol{footnote}}
\footnotetext{$^*$ Work done during internship at Tencent.}
\footnotetext{$^\dagger$ Corresponding Authors.}
\renewcommand*{\thefootnote}{\arabic{footnote}}

\begin{abstract}
On-policy distillation (OPD) transfers reasoning behavior by training a student on teacher feedback along student-generated trajectories, but standard full-rollout training ties every update to a costly completion and can over-allocate supervision to late positions with low marginal value for the current student.
We revisit this assumption through the useful supervision horizon: student-induced rollouts can drift from teacher-preferred continuations, while aligned prefixes may already preserve the long-horizon OPD update direction.
We propose ADWIN, an adaptive-window framework for OPD that treats rollout length as an online admissibility decision, training on short teacher-anchored prefixes while using delayed full-rollout probes to audit prefix--full alignment and adapt the next horizon with staleness control.
Across math and code reasoning benchmarks in single-task, multi-task, and strong-to-weak settings, ADWIN improves the accuracy--compute trade-off over full-rollout OPD and prefix-based baselines, reducing end-to-end training cost by up to $4.1\times$ while achieving comparable or better accuracy.
\end{abstract}

\newcommand{\method}{ADWIN}
\section{Introduction}

\begin{figure*}[t]
    \centering
    \includegraphics[width=1\linewidth]{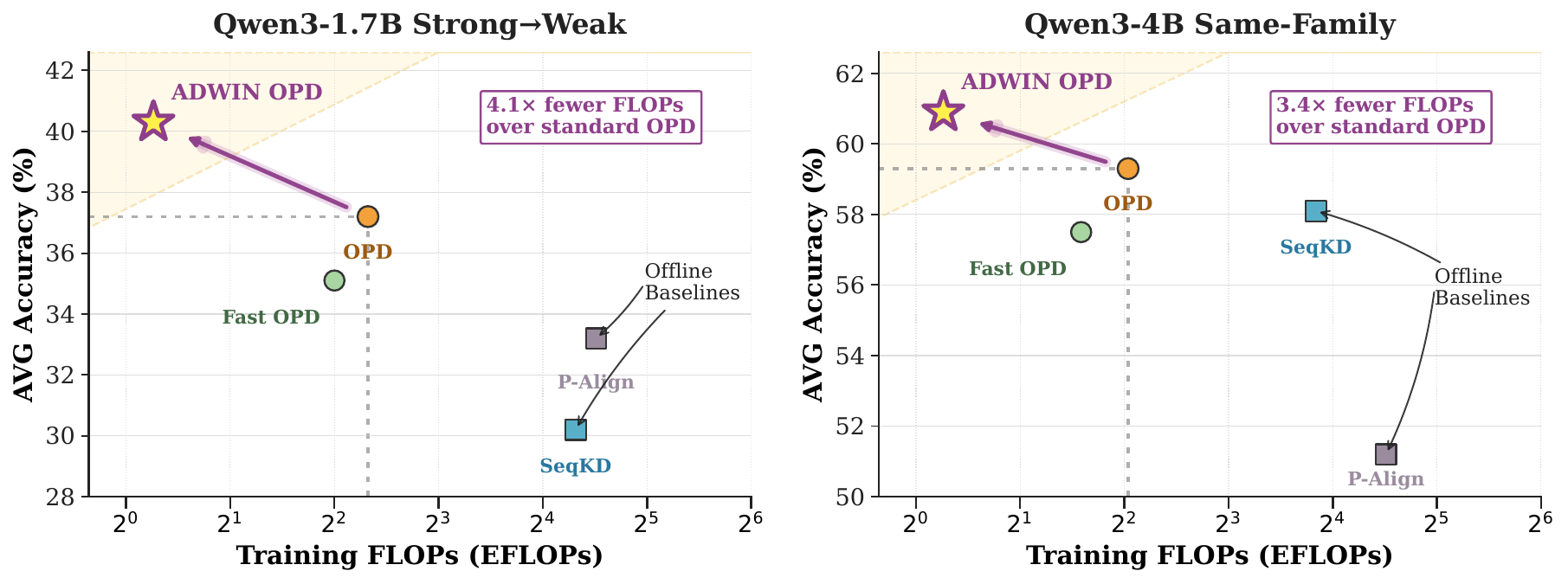}
    \caption{Accuracy--cost comparison of distillation methods. The x-axis reports log-scaled end-to-end training FLOPs and the y-axis reports average accuracy over the evaluated benchmarks. Left: Qwen3-1.7B strong-to-weak setting. Right: Qwen3-4B same-family setting, both averaged single task distillation of Math and Code tasks.}
    \label{fig:teacher_entropy}
\end{figure*}

On-policy distillation (OPD) has emerged as an important post-training paradigm for transferring reasoning behavior from a teacher model to a student model~\citep{gkd,gu2024minillm,lu2025onpolicydistillation}. 
Unlike offline knowledge distillation (KD)~\citep{kd,kim2016sequence}, which distills teacher behavior from trajectories generated by the teacher, OPD queries the teacher on trajectories sampled from the student, thereby retaining the dense supervision of distillation while grounding it in the contexts induced by the student itself. 
This on-policy nature is especially beneficial for logical reasoning, enabling supervision over the intermediate states that the student actually visits~\citep{wang-etal-2024-math}.
Recent studies have further demonstrated this potential, showing that OPD can reproduce strong reasoning gains at substantially lower compute, especially on reasoning-heavy tasks~\citep{qwen3, xiaomi2025mimo,glm5team2026glm5vibecodingagentic}.

Despite the effectiveness of this paradigm, making supervision on-policy does not by itself determine how dense sequential signal should be used during training.
Standard OPD practices take the direct operationalization, updating teacher feedback over a complete rollout. 
However, this fine-grained feedback does not imply that supervision at all positions is equally useful: in autoregressive generation, the student-induced context evolves recursively, and early decisions can causally constrain or distort the subsequent reasoning path~\citep{errorproporgation}, pushing the sequence into off-manifold states that resemble a teacher-side exposure bias~\citep{exposurebias,bengio2015scheduled}. 

The same causal structure also motivates a constructive sufficiency question for prefixes: whether an early rollout region can already provide sufficient directional evidence for the long-horizon OPD update. The joint perspective reframes OPD training window as a prefix-admissibility problem, highlighting a degree of freedom left implicit in standard OPD: how far dense teacher feedback should be instantiated along a student-induced rollout.
A further operational question follows: given the inherent rollout cost of on-policy training, can OPD identify supervision that would ultimately be unnecessary and truncate the rollout online, before the corresponding full trajectory and backward signal are available?

To this end, we propose \method{}, an \textbf{AD}aptive-\textbf{WIN}dow OPD framework that treats rollout truncation as a prefix-admissibility decision. 
Instead of tying every update to a complete trajectory, \method{} performs synchronous OPD updates on short teacher-anchored prefix windows whose gradients are intended to preserve the long-horizon OPD direction. 
In parallel, it asynchronously extends a small subset of unfinished prefixes into delayed full-rollout probes, keeping long-horizon evidence off the optimization critical path. When probes return, \method{} uses their prefix--full consistency to audit candidate prefix window sizes and calibrate the next synchronous horizon. 
This allows \method{} to select the shortest admissible window, reducing rollout cost and making updates non-blocking with respect to full-rollout completion, while still retaining long-horizon evidence for adaptive supervision.

As shown in Figure~1, ADWIN improves the OPD accuracy--cost trade-off through adaptive prefix-window updates with delayed full-rollout auditing, reducing end-to-end training FLOPs by up to 4.1$\times$ while achieving comparable or better accuracy than standard OPD and outperforming fixed or externally scheduled prefix baselines.

Our contributions are threefold:
\begin{enumerate}
    \item We formulate prefix-window selection in OPD as a prefix-admissibility problem, linking efficient online rollout truncation to the preservation of long-horizon OPD update directions.
    
    \item We introduce \method{}, an adaptive-window OPD framework that combines synchronous prefix updates with delayed full-rollout probes to audit prefix--full alignment and calibrate the training window online.
    
    \item We evaluate \method{} on math and code benchmarks across single-task, multi-task, and strong-to-weak settings, showing that it substantially reduces rollout cost while preserving or improving OPD accuracy.
\end{enumerate}

\section{Preliminaries}
\label{sec:pre}

\subsection{Notation}

Let $D$ denote a prompt distribution. Given a prompt $\boldsymbol{x}\sim D$, a response sequence is $\boldsymbol{y}=(y_1,\ldots,y_{|\boldsymbol{y}|})$ with prefix $\boldsymbol{y}_{<t}=(y_1,\ldots,y_{t-1})$. We write $\pi_{\boldsymbol{\theta}}$ for the student policy, $\pi_{\boldsymbol{\phi}}$ for a frozen teacher, and $\mathcal{D}_{\mathrm{KL}}$ for the Kullback--Leibler divergence between two distributions over responses.

\subsection{Offline Knowledge Distillation}

KD transfers knowledge from the teacher $\pi_{\boldsymbol{\phi}}$ to the student $\pi_{\boldsymbol{\theta}}$. The general form aligns the two distributions over teacher-generated trajectories,
\begin{multline}
\mathcal{J}_{\mathrm{KD}}(\boldsymbol{\theta}) = \min_{\boldsymbol{\theta}}\; \\
\mathbb{E}_{\boldsymbol{x}\sim D,\,\boldsymbol{y}\sim\pi_{\boldsymbol{\phi}}(\cdot|\boldsymbol{x})}\!
\Big[
\mathcal{D}_{\mathrm{KL}}\!\big(
\pi_{\boldsymbol{\phi}}(\boldsymbol{y}|\boldsymbol{x})\,\|\,\pi_{\boldsymbol{\theta}}(\boldsymbol{y}|\boldsymbol{x})
\big)
\Big].
\label{eq:kd}
\end{multline}

In practice, obtaining the full output distribution of a teacher LLM is often computationally prohibitive or infeasible, Eq.~\eqref{eq:kd} is commonly approximated via supervised fine-tuning (SFT) on teacher-generated trajectories. Either way, this \emph{off-policy} paradigm fails to align the student's actions with its self-induced reasoning context; by imitating fixed oracle paths, the model circumvents the causal feedback of its own decisions, leading to logical drift when navigating its own prefixes at inference time~\citep{pmlr-v15-ross11a}.

\subsection{On-Policy Distillation}

OPD~\citep{gkd,lu2025onpolicydistillation} resolves this mismatch by querying the teacher on student-generated rollouts. 
\begin{multline}
    \mathcal{J}_{\mathrm{OPD}}(\boldsymbol{\theta}) = \min_{\boldsymbol{\theta}}\; \\
    \mathbb{E}_{\boldsymbol{x}\sim D,\,\boldsymbol{y}\sim\pi_{\boldsymbol{\theta}}(\cdot|\boldsymbol{x})}\!
    \Big[
    \mathcal{D}_{\mathrm{KL}}\!\big(
        \pi_{\boldsymbol{\theta}}(\boldsymbol{y}|\boldsymbol{x})\,\|\,\pi_{\boldsymbol{\phi}}(\boldsymbol{y}|\boldsymbol{x})
    \big)
    \Big].
    \label{eq:opd}
\end{multline}

\textbf{}

Where $\boldsymbol{y}$ is sampled from the current student, making the training on-policy. Let $c_t = \log\pi_{\boldsymbol{\theta}}(y_t|\boldsymbol{x}, \boldsymbol{y}_{<t}) - \log\pi_{\boldsymbol{\phi}}(y_t|\boldsymbol{x}, \boldsymbol{y}_{<t})$ denote the step-wise cost. The standard policy gradient couples the parameter update at step $t$ with all future costs in a return-to-go fashion~\citep{schulman2018highdimensionalcontinuouscontrolusing}:

\begin{equation}
    \resizebox{1\linewidth}{!}{$
    \displaystyle
    \nabla_{\boldsymbol{\theta}} \mathcal{J}_{\mathrm{OPD}} \approx \mathbb{E}\!
    \left[
    \sum_{t=1}^{|\boldsymbol{y}|} \nabla_{\boldsymbol{\theta}} \log\pi_{\boldsymbol{\theta}}(y_t|\boldsymbol{x}, \boldsymbol{y}_{<t}) \sum_{t'=t}^{|\boldsymbol{y}|} \gamma^{t'-t} c_{t'}
    \right]
    $},
    \label{eq:future_coupled_grad}
\end{equation}

where $\gamma \in [0, 1]$ is the discount factor. Following \citet{lu2025onpolicydistillation}, we apply a discount factor of $\gamma=0$, resulting corresponding policy gradient: 

\begin{equation}
    \nabla_{\boldsymbol{\theta}} \mathcal{J}_{\mathrm{OPD}}^{\gamma=0} = \mathbb{E}\!
    \left[
    \sum_{t=1}^{|\boldsymbol{y}|} c_t \cdot \nabla_{\boldsymbol{\theta}} \log\pi_{\boldsymbol{\theta}}(y_t|\boldsymbol{x}, \boldsymbol{y}_{<t}) 
    \right].
    \label{eq:no_return_to_go}
\end{equation}

The full deduction can be found in Appendix~\ref{app:opd-gradient}.
This can be viewed as a dense on-policy RL objective, where each token carries a teacher-defined local reward signal of $-\left[ \log\pi_{\boldsymbol{\theta}}(y_t|\boldsymbol{x}, \boldsymbol{y}_{<t}) - \log\pi_{\boldsymbol{\phi}}(y_t|\boldsymbol{x}, \boldsymbol{y}_{<t}) \right]$, without waiting for rollout-level returns.

\section{Admissible Prefix Windows for OPD}
\label{sec:prefix}

The two KD paradigms in Section~\ref{sec:pre} differ in how they handle the realizability of teacher supervision under the current student:

\begin{itemize}
    \item \textbf{Off-policy KD} (Eq.~\eqref{eq:kd}) makes teacher labels realizable, but not necessarily on student-realizable states.
    \item \textbf{On-policy Distillation} (Eq.~\eqref{eq:opd}) makes states student-realizable, but long rollouts can introduce cascading deviations that render later teacher feedback less actionable.
\end{itemize}

The contrast reframes the problem from trajectory source to the useful horizon of on-policy supervision: 
where along the student rollout does teacher guidance remain useful signal rather than low-value or even noisy?
We treat this as a \textbf{window-level admissibility} of locality and long-horizon fidelity: identify early student-induced region whose update remains reliable while still preserving the long-horizon OPD direction.

\subsection{Supervision Drift in On-Policy Rollouts}
\label{subsec:drift}

\begin{figure}[t]
    \centering
    \includegraphics[width=1\linewidth]{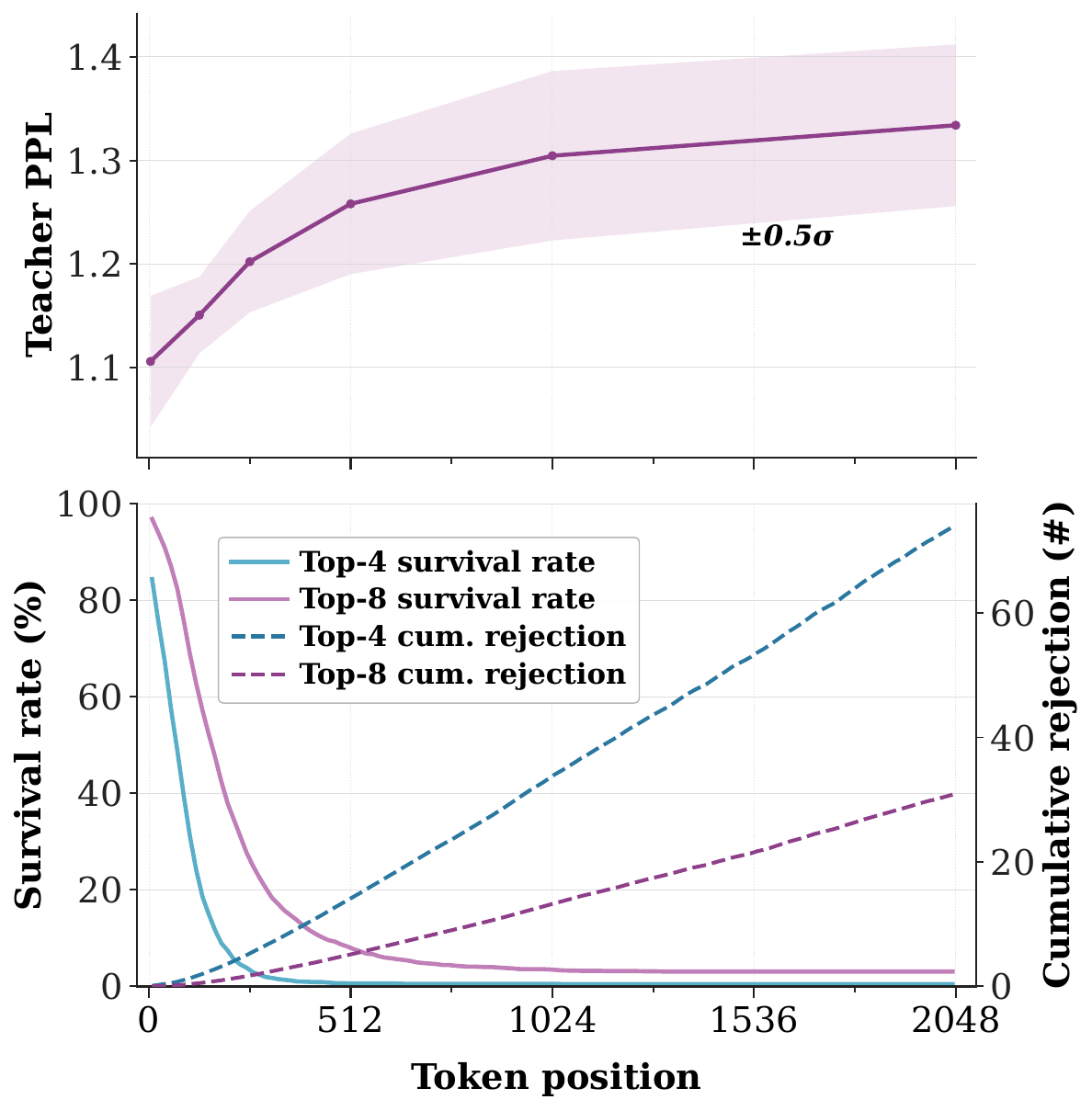}
    \caption{Per-position teacher-side drift along student-generated rollouts. Top: teacher per-position perplexity. Bottom: top-$k$ survival rate and cumulative rejection under the teacher distribution ($N=2048$ rollouts).}
    \label{fig:teacher_entropy}
\end{figure}

For a student-generated rollout $\boldsymbol{y}=(y_1,\dots,y_T)$, we track its prefix evolution under the teacher distribution and characterize the resulting drift with two diagnostics:

\begin{itemize}
    \item \textbf{Teacher branching factor:} 
    \begin{equation}
        \mathrm{BF}_t = \exp\Big(H[\pi_{\boldsymbol{\phi}}(\cdot \mid \boldsymbol{y}_{<t})]\Big),
    \end{equation}
    capturing the local multiplicity of plausible teacher paths, this quantity can be interpreted as the teacher's per-position \emph{perplexity} (PPL).
    \item \textbf{Top-$k$ rejection:} 
    We track top-$k$ departures over $N$ student-generated rollouts. 
    Let $r_{i,t}$ denote the rank of token $y_{i,t}$ under $\pi_{\boldsymbol{\phi}}(\cdot \mid \boldsymbol{y}_{i,<t})$. 
    The survival fraction up to position $T$ is
    \[
    S_T^{(k)}
    =
    \frac{1}{N}\sum_{i=1}^{N}
    \mathbbm{1}\!\left[
    \forall t \le T,\; r_{i,t} \le k
    \right],
    \]
    i.e., the fraction of rollouts with no top-$k$ rejection through position $T$.
\end{itemize}

As shown in Figure~\ref{fig:teacher_entropy}, under student-induced prefixes, the teacher distribution becomes broader at later positions, as reflected by the rising PPL. 
Meanwhile, the rapid collapse of top-$k$ survival and the steady growth of cumulative rejection show that realized rollouts increasingly escape the teacher's preferred continuations, pushing later prefixes into regions the teacher would be unlikely to reach by itself. This compounding drift induces a reversed exposure bias on the teacher side. The teacher thereby does not provide uniformly reliable supervision, which can also be potentially harmful. We provide some qualitative cases in Appendix~\ref{app:case}.

\subsection{Prefix Utility by Gradient Alignment}
\label{sec:grad_align}

Following our observations on the locality side, we further consider a training-dynamic hinged indicator to quantify prefix region's optimization utility.

\paragraph{Surrogate Descent Utility.}
Consider a local neighborhood around ${\boldsymbol{\theta}} \in {\boldsymbol{\Theta}}$. Let 
${\boldsymbol{g}} := \nabla_{\boldsymbol{\theta}} \mathcal{J}({\boldsymbol{\theta}}) \in \mathbb{R}^d$ denote the exact 
gradient of the target OPD objective, and let 
${\boldsymbol{M}}({\boldsymbol{\theta}}) \approx \mathbb{E}_{\boldsymbol{y} \sim \pi_{\boldsymbol{\theta}}} \big[ (\nabla_{\boldsymbol{\theta}} \log \pi_{\boldsymbol{\theta}}(\boldsymbol{y})) (\nabla_{\boldsymbol{\theta}} \log \pi_{\boldsymbol{\theta}}(\boldsymbol{y}))^\top \big]$ 
be a local quadratic form representing curvature. 
The second-order surrogate improvement of a parameter update ${\boldsymbol{\Delta}}$ is then given by
\begin{equation}
\mathcal{S}_{\boldsymbol{M}}({\boldsymbol{\Delta}}) := {\boldsymbol{g}}^\top {\boldsymbol{\Delta}} - \frac{1}{2}{\boldsymbol{\Delta}}^\top {\boldsymbol{M}}({\boldsymbol{\theta}}) {\boldsymbol{\Delta}}.
\end{equation}

For a surrogate update direction ${\boldsymbol{d}} \neq 0$, the maximum utility along ${\boldsymbol{d}}$ is
\begin{equation}
\mathcal{U}_{\boldsymbol{M}}({\boldsymbol{d}}) := \max_{\eta \ge 0} \mathcal{S}_{\boldsymbol{M}}(\eta {\boldsymbol{d}}) 
= \frac{[{\boldsymbol{g}}^\top {\boldsymbol{d}}]_+^2}{2 \, {\boldsymbol{d}}^\top {\boldsymbol{M}}({\boldsymbol{\theta}}) {\boldsymbol{d}}},
\label{eq:ray-utility}
\end{equation}
where $[\cdot]_+ := \max\{\cdot,0\}$. The unconstrained optimal 
update in this local geometry is the natural gradient 
${\boldsymbol{\Delta}}^\star = {\boldsymbol{M}}({\boldsymbol{\theta}})^{-1} {\boldsymbol{g}}$, which achieves the maximum local utility 
\begin{equation}
\mathcal{U}_{\boldsymbol{M}}^\star := \frac{1}{2} {\boldsymbol{g}}^\top {\boldsymbol{M}}({\boldsymbol{\theta}})^{-1} {\boldsymbol{g}}.
\end{equation}

The \emph{relative utility} of an update ${\boldsymbol{d}}$ is then
\begin{equation}
\begin{aligned}
\mathcal{E}_{\boldsymbol{M}}({\boldsymbol{d}}) := \frac{\mathcal{U}_{\boldsymbol{M}}({\boldsymbol{d}})}{\mathcal{U}_{\boldsymbol{M}}^\star} 
&= \frac{[{\boldsymbol{g}}^\top {\boldsymbol{d}}]_+^2}{({\boldsymbol{g}}^\top {\boldsymbol{M}}({\boldsymbol{\theta}})^{-1} {\boldsymbol{g}})({\boldsymbol{d}}^\top {\boldsymbol{M}}({\boldsymbol{\theta}}) {\boldsymbol{d}})} \\
&= \bigl[\cos_{\boldsymbol{M}}({\boldsymbol{d}}, {\boldsymbol{M}}({\boldsymbol{\theta}})^{-1} {\boldsymbol{g}})\bigr]_+^2.
\end{aligned}
\label{eq:utility-cos}
\end{equation}

\begin{figure*}[t]
    \centering
    \includegraphics[width=1\linewidth, height=0.53\linewidth]{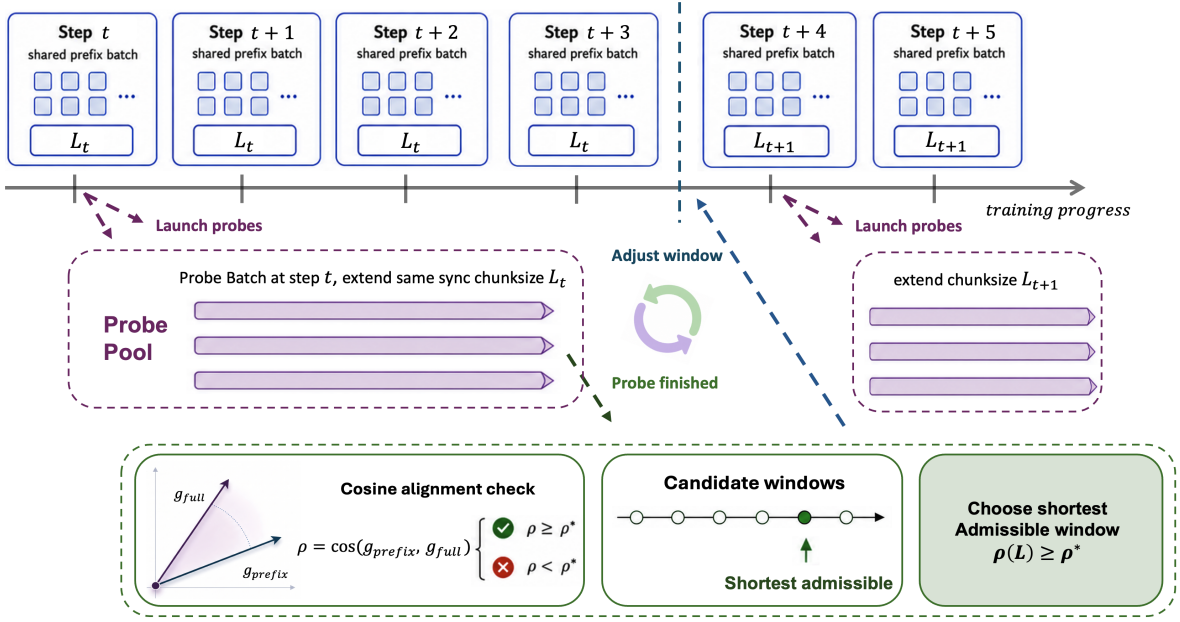}
    \caption{
    Workflow of ADWIN.
    The top row performs synchronous OPD updates on the current prefix window $L_t$: student rollouts are truncated at the shared window, teacher feedback is evaluated on the prefix, and the resulting loss is used for training updates.
    The middle probe path asynchronously continues a small probe batch of unfinished prefixes into delayed full-rollout probes.
    Returned probes are then used in the bottom audit panel to compute prefix--full gradient cosine for candidate window lengths, after which the scheduler updates $L_{t+1}$ by choosing the shortest length whose alignment exceeds the admissibility threshold $\rho^\ast$.
    }
    \label{fig:adwin_framework}
\end{figure*}

\paragraph{Evaluating Prefixes for Training Surrogates}
Equation~\eqref{eq:utility-cos} condenses the local update utility into a single interpretable scalar: the metric-aware cosine between the candidate update and the ideal natural-gradient direction \({\boldsymbol{M}}({\boldsymbol{\theta}})^{-1}{\boldsymbol{g}}\). For a prefix gradient \({\boldsymbol{d}}_p\) and a full-rollout gradient \({\boldsymbol{d}}_f\), their relative informational value is
\begin{equation}
\frac{\mathcal{E}_{\boldsymbol{M}}({\boldsymbol{d}}_p)}{\mathcal{E}_{\boldsymbol{M}}({\boldsymbol{d}}_f)}
=
\left(
\frac{\cos_{\boldsymbol{M}}\bigl({\boldsymbol{d}}_p,{\boldsymbol{M}}({\boldsymbol{\theta}})^{-1}{\boldsymbol{g}}\bigr)}
     {\cos_{\boldsymbol{M}}\bigl({\boldsymbol{d}}_f,{\boldsymbol{M}}({\boldsymbol{\theta}})^{-1}{\boldsymbol{g}}\bigr)}
\right)^2 .
\label{eq:prefix_vs_full_ratio}
\end{equation}

A prefix is therefore an effective surrogate if its gradient preserves the optimization-relevant \(M\)-cosine. In the transparent case \({\boldsymbol{d}}_f={\boldsymbol{M}}({\boldsymbol{\theta}})^{-1}{\boldsymbol{g}}\), Eq.~\eqref{eq:utility-cos} reduces to \(\mathcal{E}_{\boldsymbol{M}}({\boldsymbol{d}}_p)=\cos_{\boldsymbol{M}}^2({\boldsymbol{d}}_p,{\boldsymbol{d}}_f)\). This simplification provides an operational pathway for practically scheduling the training window.

\section{ADWIN: Pushing Acceleration to Online Rollout Truncation}

Although a prefix cannot give an unbiased estimate of the long-horizon update direction, Section~\ref{sec:prefix} shows it can still yield a useful descent step on the OPD objective. 
We further exploit it as an operational question: given that window selection itself requires long-horizon evidence, can OPD adaptively truncate rollouts online without first incurring the full-generation cost that motivates truncation?
To remedy this, we propose ADWIN, an adaptive-window training framework for OPD that makes region-size selection online and auditable, using delayed asynchronous probes to adapt the horizon of synchronous truncated updates.

\subsection{Sync Updates and Async Probes}

ADWIN separates OPD training into a synchronous update path and an asynchronous audit path. 
At optimization step $t$, the synchronous path uses only the current window length $L_t$: it samples student continuations up to $L_t$, evaluates the teacher-anchored OPD loss on this prefix, and immediately applies the resulting update. 
Thus the blocking cost of a training step is
$C_{\mathrm{sync}}(t)
=
C_{\mathrm{roll}}(L_t)+C_{\mathrm{sup}}(L_t),
$
rather than the full-rollout cost
$C_{\mathrm{roll}}(L_{\max})+C_{\mathrm{sup}}(L_{\max})$.

The full horizon is accessed only through delayed probes. 
After each synchronous batch is formed, ADWIN selects a small subset of unfinished prefixes and inserts them into a probe pool $\mathcal{P}$. 
Background workers continue these probes toward $L_{\max}$ over subsequent training steps, with each extension round capped by the current synchronous rollout budget. 
Consequently, full rollouts are not eliminated, but moved off the optimization critical path: they are used only to produce delayed audit batches for future window calibration.

To control the mismatch between delayed probes and the evolving student policy, ADWIN enforces a hard staleness limit measured in optimization steps. 
A probe that exceeds this limit is forced to complete before being consumed by the scheduler. 
The returned probe batch is then passed to the admissibility audit described below.

\subsection{Probe-Based Window Calibration}

Upon a probe's return, we evaluate a candidate list of truncation lengths $\mathcal{L} = \{L_1, L_2, \dots\}$ to update the synchronous window $L_t$. Both the candidate prefix gradient $\bar u_t^{(L)}$ and the full-horizon probe reference $u_t^{\mathrm{probe}}$ are evaluated via a forward pass under the current student parameters. Following the geometric criteria established in Section~\ref{sec:grad_align}, we compute their metric-aware consensus $\rho_t(L) := \cos\!\big(u_t^{\mathrm{probe}}, \bar u_t^{(L)}\big)$.

Decomposing $u_t^{\mathrm{probe}}$ into an aligned component and an orthogonal residual yields the prefix surrogate's effective signal-to-noise ratio: $\mathrm{SNR}_t(L) := \rho_t(L)^2 / (1 - \rho_t(L)^2)$. Requiring a strictly prefix signal dominated update ($\mathrm{SNR}_t(L) \ge 1$) as a uniform approximation directly derives our empirical lower bound admissibility threshold:
\begin{equation}
\rho_t(L) \ge \rho^* = \frac{\sqrt{2}}{2} \approx 0.707.
\label{eq:snr-threshold}
\end{equation}

The scheduler updates the online training horizon to the shortest length that satisfies this alignment bound: $L_{t+1} = \min \{ L \in \mathcal{L} \mid \rho_t(L) \ge \rho^* \}$. 
If the current prefix demonstrates strong gradient consensus ($\ge\rho^*$), the tracking window shrinks to maximize wall-clock acceleration; conversely, if the alignment drops below the conservative signal-to-noise threshold, the window gracefully expands to reclaim supervision fidelity. 

\section{Experiments}

To comprehensively evaluate ADWIN, we conduct experiments on math and code generation tasks across three distinct settings: single-task, multi-task, and strong-to-weak.

\subsection{Experimental Setup}
\begin{table*}[!t]
\centering
\small
\renewcommand{\arraystretch}{1.1}
\setlength{\tabcolsep}{5pt}
\begin{tabular}{lcccccccc}
\toprule
\multirow{3}{*}{\textbf{Method}}
& \multicolumn{3}{c}{\textbf{Mathematics}}
& \multicolumn{3}{c}{\textbf{Code}}
& \multirow{3}{*}{\textbf{Avg.}}
& \multirow{3}{*}{\textbf{EFLOPs $\downarrow$}} \\
\cmidrule(lr){2-4}\cmidrule(lr){5-7}
& AIME'24 & AIME'25 & Beyond AIME & HumanEval+ & MBPP+ & LCB v6 & & \\
& \texttt{\footnotesize mean@16} & \texttt{\footnotesize mean@16} & \texttt{\footnotesize mean@8}
& \texttt{\footnotesize mean@4}  & \texttt{\footnotesize mean@4}  & \texttt{\footnotesize mean@4} & & \\
\midrule
Initial Model & 22.5 & 18.9 & 10.8 & 71.9 & 60.8 & 24.6 & 34.9 & -- \\
Teacher       & 61.3 & 58.9 & 34.0 & 83.4 & 75.6 & 32.0 & 57.5 & -- \\
\midrule
\multicolumn{9}{l}{\emph{\textbf{(a) Single-Task}}} \\
\rowcolor{gray!6} SeqKD   & 61.7 & 56.9 & 34.0 & 82.2 & 75.5 & 38.0 & 58.1 & 14.3 \\
\rowcolor{gray!6} P-Align & 54.6 & 52.9 & 29.8 & 77.8 & 63.0 & 29.3 & 51.2 & 22.8 \\
OPD             & 58.8 & 58.5 & 32.9 & \textbf{92.6} & 75.9 & 37.1 & 59.3 & \phantom{0}4.1 \\
Fast OPD        & 60.4 & 54.9 & 33.3 & 83.3 & 76.0 & 37.3 & 57.5 & \phantom{0}3.0 \\
\rowcolor{cyan!10} ADWIN & \textbf{65.6} & \textbf{62.9} & \textbf{34.9} & 85.9 & \textbf{77.4} & \textbf{38.6} & \textbf{60.9} & \phantom{0}\textbf{1.2} \\
\midrule
\multicolumn{9}{l}{\emph{\textbf{(b) Multi-Task}}} \\
\rowcolor{gray!6} SeqKD   & 61.0 & 55.2 & 33.3 & 84.2 & 75.7 & 38.1 & 57.9 & 25.7 \\
\rowcolor{gray!6} P-Align & 53.5 & 50.8 & 31.8 & 77.8 & 65.4 & 31.6 & 51.8 & 46.7 \\
OPD             & 60.4 & 56.1 & \textbf{34.8} & 83.8 & 77.0 & 37.1 & 58.2 & \phantom{0}8.2 \\
Fast OPD        & 62.1 & 57.9 & 33.9 & 83.1 & 76.2 & 37.9 & 58.5 & \phantom{0}6.5 \\
\rowcolor{cyan!10} ADWIN & \textbf{64.2} & \textbf{59.8} & 34.6 & \textbf{84.9} & \textbf{77.4} & \textbf{38.4} & \textbf{59.9} & \phantom{0}\textbf{2.7} \\
\bottomrule
\end{tabular}
\caption{Main results in the same-family settings. Student is Qwen3-4B-Non-Thinking; teachers are the corresponding RL-tuned checkpoints. Gray rows mark offline-distillation baselines. \textbf{Bold} marks the best student method within each setting; the highlighted row is ADWIN.}
\label{tab:peer}
\end{table*}

\begin{table*}[!t]
    \centering
    \small
    \renewcommand{\arraystretch}{1.1}
    \setlength{\tabcolsep}{5pt}
    \begin{tabular}{lcccccccc}
    \toprule
    \multirow{3}{*}{\textbf{Method}}
    & \multicolumn{3}{c}{\textbf{Mathematics}}
    & \multicolumn{3}{c}{\textbf{Code}}
    & \multirow{3}{*}{\textbf{Avg.}}
    & \multirow{3}{*}{\textbf{EFLOPs $\downarrow$}} \\
    \cmidrule(lr){2-4}\cmidrule(lr){5-7}
    & AIME'24 & AIME'25 & Beyond AIME & HumanEval+ & MBPP+ & LCB v6 & & \\
    & \texttt{\footnotesize mean@16} & \texttt{\footnotesize mean@16} & \texttt{\footnotesize mean@8}
    & \texttt{\footnotesize mean@4}  & \texttt{\footnotesize mean@4}  & \texttt{\footnotesize mean@4} & & \\
    \midrule
    Initial Model & 13.3 & \phantom{0}7.9 & \phantom{0}4.5 & 57.1 & 40.2 & 13.1 & 22.7 & -- \\
    Teacher       & 76.0 & 61.9           & 44.8           & 92.6 & 79.4 & 40.9 & 65.9 & -- \\
    \midrule
    \multicolumn{9}{l}{\emph{\textbf{(c) Strong-to-Weak}}} \\
    \rowcolor{gray!6} SeqKD   & 19.0 & 16.5 & \phantom{0}7.6 & 65.3 & 59.5 & 13.1 & 30.2 & 19.9 \\
    \rowcolor{gray!6} P-Align & 27.1 & 19.6 & \phantom{0}8.8 & 64.7 & 60.9 & 18.1 & 33.2 & 22.8 \\
    OPD             & 30.4 & 22.5 & 10.8 & \textbf{72.7} & 64.0 & 22.6 & 37.2 & \phantom{0}5.0 \\
    Fast OPD        & 32.1 & \textbf{27.7} & 12.8 & 62.1 & 57.7 & 18.0 & 35.1 & \phantom{0}4.0 \\
    \rowcolor{cyan!10} ADWIN & \textbf{34.2} & 26.9 & \textbf{13.4} & 72.4 & \textbf{67.1} & \textbf{27.7} & \textbf{40.3} & \phantom{0}\textbf{1.2} \\
    \bottomrule
    \end{tabular}
    \caption{Main results in the strong-to-weak setting. The student is Qwen3-1.7B-Non-Thinking, supervised by the stronger Qwen3-30B-A3B-Instruct-2507.}   
    \label{tab:s2w}
\end{table*}

\paragraph{Base Models}
For the same-family single- and multi-task settings, we use Qwen3-4B-Non-Thinking~\citep{qwen3} as the student and supervise it with the corresponding RL-tuned experts. For the strong-to-weak setting, we use Qwen3-1.7B-Non-Thinking as the student and Qwen3-30B-A3B-Instruct-2507 as the stronger teacher.

\paragraph{Training Details.}
We train the math task on Polaris-53K~\citep{Polaris2025}, and the code task on Eurus-RL-Code~\citep{euruscode} for both RL and OPD phases. The RL experts are obtained with GRPO~\citep{shao2024deepseekmath} using a binary correctness reward, with 128 prompts per batch and 8 rollouts. We implement ADWIN with candidate training windows $\mathcal{L}=\{64, 128, 256, 512, 1024, 2048\}$ under a maximum training horizon $L_{\max}$ of 8,192 tokens, selecting the shortest window whose probe-estimated prefix--full gradient cosine exceeds $\rho^*$; the probe batch size is set to 64 samples and maximum probe staleness is set to 5 optimization steps. See Appendix~\ref{app:train_details} for more details.

\paragraph{Baselines.}
We evaluate ADWIN against different prefix-level designs of distillation. For OPD-family baselines, standard OPD updates on complete student trajectories; Fast OPD~\citep{zhang2026fast} uses a predefined schedule that linearly increases the prefix window; and fixed-window Prefix OPD trains with manually chosen prefixes without adaptive auditing. For offline baselines, SeqKD~\citep{kim2016sequence} trains on teacher-generated full trajectories, while P-Align~\citep{liu2026long} truncates teacher trajectories at a prefix length judged sufficient by the student, who then generates the remaining suffix for continuation. All baselines use the same student, teacher; specific implementation details can be found in Appendix~\ref{app:baseline_details}. 

\paragraph{Evaluations Setup}
We evaluate on three challenging mathematical reasoning tasks: AIME 2024~\citep{aime2024} AIME 2025~\citep{aime2025}, BeyondAIME~\citep{bytedance_seed_2025_beyondaime} and three coding tasks: HumanEval+, MBPP+~\citep{evalplus}, LivecodeBench (v6 only)~\citep{livecodebench}
. In all the evaluations, we set the temperature to 0.6, top-p to 0.95, top-k to 20 and maximum generation length to 30,720 tokens. 

\subsection{Main Results}
\paragraph{ADWIN improves the end-to-end accuracy--cost trade-off of OPD.}
Tables~\ref{tab:peer} and~\ref{tab:s2w} report the main results across the single-RL-teacher, multitask, and strong-to-weak settings. 
ADWIN achieves the best average performance among all evaluated distillation methods in all three settings while using substantially lower end-to-end compute than standard OPD.
Compared with full-rollout OPD, ADWIN improves the average score from 59.3 to 60.9 in the single-RL-teacher setting, from 58.2 to 59.9 in the multitask setting, and from 37.2 to 40.3 in the strong-to-weak setting. 
At the same time, ADWIN reduces EFLOPs from 4.1 to 1.2, 8.2 to 2.7, and 5.0 to 1.2, respectively, corresponding to up to a 4.1$\times$ reduction in end-to-end training cost.

\paragraph{Prefix truncation needs dynamic admissibility.}
ADWIN also improves over two prefix-level designs, P-Align and Fast OPD, where prefixes are selected either by offline student self-evaluation or by an externally predefined schedule. This supports the advantage of internalizing prefix selection into the OPD training dynamics: the effective horizon should be treated not as a fixed preprocessing decision, but as a training-time control variable responsive to evolving student--teacher interactions.

\subsection{Analysis and Ablations}
For all analysis and ablations, we adopt the more discriminative strong-to-weak setting to provide a clear contrast, while the results under the peer distillation setup follow the same trajectory.

\paragraph{Effect of adaptive window selection.}

\begin{figure*}
    \centering
    \includegraphics[width=1.0\linewidth]{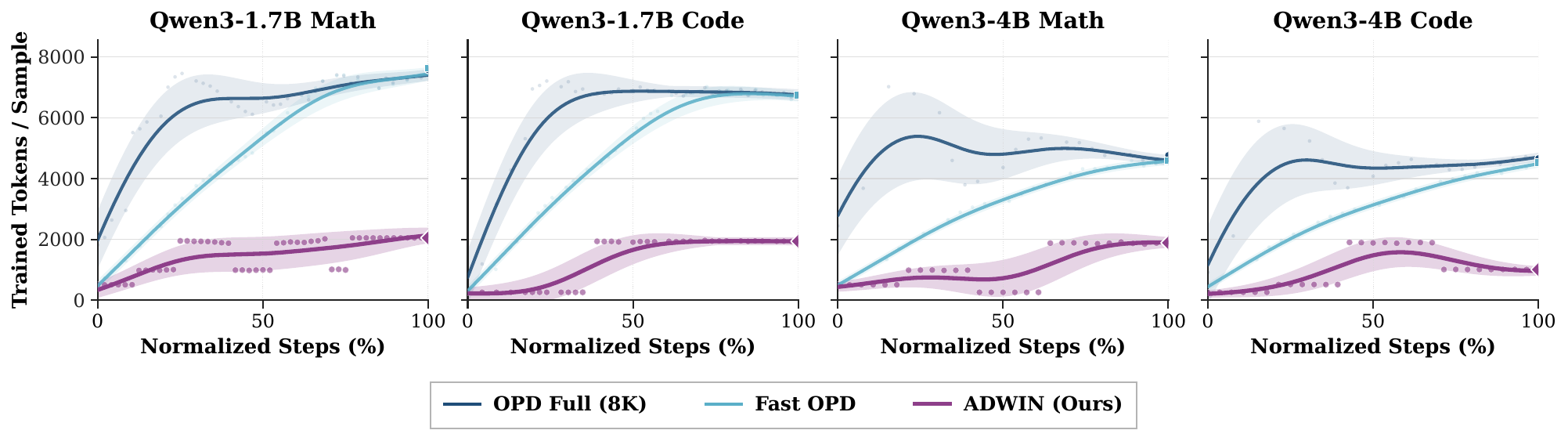}
    \caption{Evolution of the effective training horizon during OPD. 
    The curves report per step trained tokens over normalized training progress under full-rollout OPD, Fast OPD, and ADWIN. ADWIN maintains a substantially smaller horizon than full-rollout OPD and externally scheduled truncation by adapting the prefix window online.}
\label{fig:token_dynamics}
\end{figure*}
 We ablate the performance by replacing ADWIN's scheduler with fixed-window Prefix candidates used in training. As shown in Table~\ref{tab:fixed_window_ablation}, no static horizon is uniformly reliable: overly short windows can miss necessary long-horizon signal, while longer windows do not monotonically improve performance and may reintroduce low-value late-position supervision. This indicates that the useful horizon is not a tunable length fixed before training, but a training-time admissibility boundary that varies across settings and hinges on training dynamics.

\begin{table}[!t]
\centering
\small
\setlength{\tabcolsep}{3pt}
\begin{tabular}{lccc}
\toprule
\textbf{Window policy} & \textbf{Math Avg.} & \textbf{Code Avg.} & \textbf{Overall Avg.} \\
\midrule
Prefix OPD-128 & 15.2 & 50.3 & 32.8 \\
Prefix OPD-256 & 21.3 & 49.5 & 35.4 \\
Prefix OPD-512 & 23.9 & 54.7 & 39.3 \\
Prefix OPD-1024 & 24.0 & \textbf{56.2} & 40.1 \\
Prefix OPD-2048 & 24.0 & 52.5 & 38.3 \\
\rowcolor{cyan!10} ADWIN & \textbf{24.8} & 55.7 & \textbf{40.3} \\
\bottomrule
\end{tabular}
\caption{Strong-to-weak fixed-window ablation. Prefix OPD-$L$ trains with a $L$-token window. Math/Code Avg. average the three benchmarks within each domain.}
\label{tab:fixed_window_ablation}
\end{table}

\paragraph{Reliable gradient consensus enables window selection.}

We next examine an empirical observation that motivates the design of ADWIN: prefix gradients can form a reliable consensus with the full-rollout OPD gradient after batch aggregation.

\paragraph{The cross-sample consensus.}

\begin{figure*}[t]
  \includegraphics[width=0.48\linewidth]{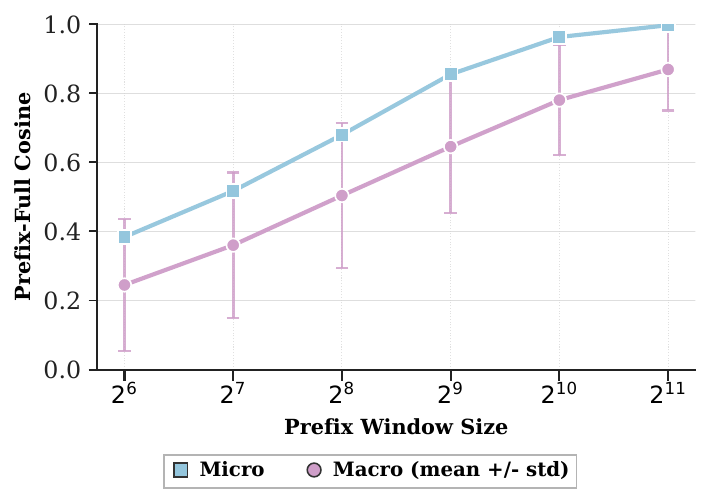} \hfill
  \includegraphics[width=0.48\linewidth]{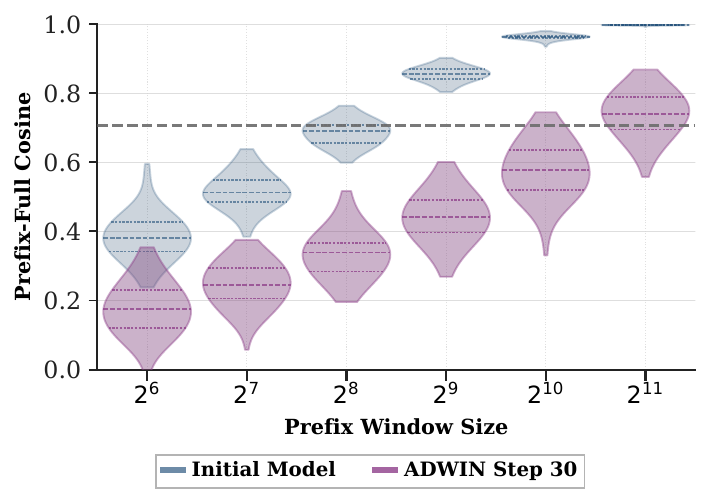}
\caption{
Prefix--full gradient cosine analysis on Qwen3-1.7B strong-to-weak setting, evaluated on Polaris.
Left: macro cosine averages per-sample prefix--full alignments, while micro cosine is computed after batch-level gradient aggregation..
Right: distributions of batch-level micro prefix--full cosine across 128 independent probe batches at the initial model and a middle-stage ADWIN checkpoint.
}
\label{fig:gradient_consensus}
\end{figure*}

We first study the consensus effect of per-sample gradient aggregation, as illustrated in Figure~\ref{fig:gradient_consensus}. The batch-level micro prefix--full cosine rises quickly at short prefix lengths, even when the prefix covers only a small fraction of the full rollout. The macro cosine, obtained by averaging per-sample prefix--full alignments, remains consistently lower. This suggests a cancellation over high variance components across samples, which makes the prefix updates more reliable and useful signal even at the early positions, echoed by the non-trivial performance of aggressive truncated settings in Table~\ref{tab:fixed_window_ablation}.

\paragraph{The cross-batch consensus.} 
We then examine whether the prefix-full gradient structure holds across batches, since which ensures the reliability of window selection in ADWIN's gradient probe.
As shown in Figure~\ref{fig:gradient_consensus}, the gradient cosine remains highly stable across various probe batches. Consequently, the distribution across the threshold that triggers window decision flips remains minimal. Most importantly, we observe no cross-tier flips: sub-threshold cases do not simultaneously appear across non-adjacent window tiers, supporting the robustness of our gradient probe.

\section{Related Work}

\paragraph{Knowledge distillation for reasoning.} We broadly categorize KD works by where the training data are sampled, resulting in offline~\citep{kd,kim2016sequence} and on-policy paradigms~\citep{gkd,lu2025onpolicydistillation},  yields different KL choices rooted in the inherent estimation principles of KL divergence, as well as subsequent works that optimize the use of these losses(e.g. adaptively combining these two)\citep{wu2024rethinkingkullbackleiblerdivergenceknowledge,wang2025abkd}. Within the focus of OPD, recent advances further push broader applications including: optimization stability~\citep{jang2026stable, fu2026revisiting}, cross-family transfer~\citep{minixhofer2025universalcrosstokenizerdistillationapproximate}, self-distillation~\citep{self-distilled-reasoning,SDPO}, and distillation-to-RL extension~\citep{gopd,rlad,rlsd}.
ADWIN focuses on a complementary question: selecting sufficient prefix windows for efficient OPD training.

\paragraph{Selective supervision for distillation.} Efficient dense supervision often reduces redundant token-level updates by masking, filtering, or reweighting losses according to confidence, uncertainty, teacher--student disagreement, or estimated utility~\citep{huang2025selectkd,todi,sang2026not,kim2026explainwordsimprovingreasoning}. 
Such methods improve how generated tokens are used, but typically do not reduce the rollout generation cost that dominates OPD training. 
Fast OPD~\citep{zhang2026fast} directly reduces this cost through externally scheduled prefix truncation, but its horizon is not audited against the current full-rollout update. 
ADWIN instead makes truncation an online admissibility decision: it trains on short prefix updates while using delayed probes to verify their alignment with the full-rollout OPD direction.

\section{Conclusion}
This paper presents ADWIN, an adaptive-window framework for efficient on-policy distillation. 
We show that full-rollout supervision is not uniformly necessary: student-induced drift can reduce the value of late-token feedback, while aligned prefixes can still preserve the long-horizon OPD update direction. 
ADWIN turns rollout truncation into a window-admissibility problem, training on short teacher-anchored prefixes while using delayed full-rollout probes to audit prefix--full gradient alignment and adapt the synchronous horizon. 
Across math and code benchmarks in single-task, multi-task, and strong-to-weak settings, ADWIN improves the accuracy--compute trade-off over full-rollout OPD and prefix-based baselines, substantially reducing end-to-end training cost.

\section*{Limitations}
We elucidate the limitations of this work as follows.
First, ADWIN adopts a synchronized truncation window for each training update rather than sample-wise length selection, in order to keep the online training path regular and reduce straggler-induced bubbles caused by heterogeneous rollout lengths within a batch. Consequently, the admissibility decision is made through batch-level prefix--full gradient consensus rather than per-example sufficiency. Although the threshold $\rho^\ast$ provides a practical criterion for preserving the aggregated update direction and the empirical results are favorable, it does not theoretically guarantee that synchronized truncation preserves the performance of full-rollout OPD in all cases. Second, this work focuses on reasoning-intensive tasks, while the generalization of ADWIN to open-ended dialogue, instruction following, or multimodal reasoning remains untested. Third, ADWIN may be less suitable for tasks whose decisive supervision signals appear primarily near the end of a trajectory. Prefix--full gradient alignment mainly characterizes the consistency of optimization directions, and may not fully reflect suffix-specific requirements such as final-answer verification, proof closure, code executability, or long-range dependency repair.

\section*{Ethical Considerations}

\subsection*{Use of AI Assistants}

The algorithmic design and main methodology presented in this work were developed through human-led research and reasoning. During implementation, we utilized GitHub Copilot 3\footnote{https://github.com/features/copilot} for coding assistance. We affirm that the primary content and logic of the code are entirely our own work.
\subsection*{Potential Risks}

This work introduces ADWIN (ADaptive WINdows), a framework for efficient on-policy distillation through adaptive rollout truncation. We acknowledge potential risks associated with accelerating supervision for reasoning models. Distilled students may inherit incorrect reasoning patterns, biases, or unsafe behaviors from the teacher and training data. Moreover, by reducing the cost of on-policy distillation, ADWIN may make it easier to scale the training and deployment of distilled reasoning models before their failure modes are fully understood. If used improperly, such efficiency gains could amplify teacher errors, dataset biases, or brittle reasoning behaviors across larger numbers of downstream models and applications.

\bibliography{custom}

\appendix
\label{sec:app}

\section{Full Derivation of OPD and Descent Utility}

\subsection{Deriving the Token-Level OPD Gradient}
\label{app:opd-gradient}

We derive the token-level estimator used in Section~\ref{sec:pre}.
For $\boldsymbol{x}\sim\mathcal{D}$, let $\boldsymbol{y}=(y_1,\ldots,y_T)\sim\pi_{\boldsymbol{\theta}}(\cdot|\boldsymbol{x})$ and $\boldsymbol{s}_t=(\boldsymbol{x},\boldsymbol{y}_{<t})$. For compactness, define the token-level gradient $\boldsymbol{g}_t$ and the teacher-relative cost $c_t$ as:
\begin{align}
\boldsymbol{g}_t &:= \nabla_{\boldsymbol{\theta}} \log\pi_{\boldsymbol{\theta}}(y_t|\boldsymbol{s}_t), \\
c_t &:= \log\pi_{\boldsymbol{\theta}}(y_t|\boldsymbol{s}_t) - \log\pi_{\boldsymbol{\phi}}(y_t|\boldsymbol{s}_t).
\end{align}
By autoregressive factorization, the sequence-level differences decompose as:
\begin{align}
\log\frac{\pi_{\boldsymbol{\theta}}(\boldsymbol{y}|\boldsymbol{x})}{\pi_{\boldsymbol{\phi}}(\boldsymbol{y}|\boldsymbol{x})} &= \sum_{t=1}^T c_t, \\
\nabla_{\boldsymbol{\theta}}\log\pi_{\boldsymbol{\theta}}(\boldsymbol{y}|\boldsymbol{x}) &= \sum_{t=1}^T \boldsymbol{g}_t.
\end{align}

The sequence-level OPD objective can therefore be written as:
\begin{equation}
J_{\mathrm{OPD}}(\boldsymbol{\theta}) = \mathbb E_{\boldsymbol{x},\boldsymbol{y}} \bigg[ \sum_{t=1}^T c_t \bigg], \qquad \boldsymbol{y}\sim\pi_{\boldsymbol{\theta}}(\cdot|\boldsymbol{x}).
\end{equation}

Applying the score-function identity directly yields:
\begin{equation}
\begin{aligned}
\nabla_{\boldsymbol{\theta}} J_{\mathrm{OPD}}(\boldsymbol{\theta})
&= \mathbb E_{\boldsymbol{x},\boldsymbol{y}} \bigg[ \Big( \sum_{t'=1}^T c_{t'} \Big) \Big( \sum_{t=1}^T \boldsymbol{g}_t \Big) \bigg] \\
&\quad + \mathbb E_{\boldsymbol{x},\boldsymbol{y}} \bigg[ \nabla_{\boldsymbol{\theta}} \sum_{t'=1}^T c_{t'} \bigg].
\end{aligned}
\end{equation}
Since the teacher is fixed, $\nabla_{\boldsymbol{\theta}} c_{t'} = \nabla_{\boldsymbol{\theta}} \log\pi_{\boldsymbol{\theta}}(y_{t'}|\boldsymbol{s}_{t'}) = \boldsymbol{g}_{t'}$. The second term vanishes in expectation because the expected score is zero:
\begin{equation}
\mathbb E_{\boldsymbol{x},\boldsymbol{y}} \bigg[ \sum_{t'=1}^T \boldsymbol{g}_{t'} \bigg] = \mathbb E_{\boldsymbol{x},\boldsymbol{y}} [ \nabla_{\boldsymbol{\theta}}\log\pi_{\boldsymbol{\theta}}(\boldsymbol{y}|\boldsymbol{x}) ] = \mathbf{0}.
\end{equation}
Thus, the gradient simplifies to the cross-terms:
\begin{equation}
\nabla_{\boldsymbol{\theta}} J_{\mathrm{OPD}}(\boldsymbol{\theta}) = \mathbb E_{\boldsymbol{x},\boldsymbol{y}} \bigg[ \sum_{t=1}^T \sum_{t'=1}^T c_{t'} \boldsymbol{g}_t \bigg].
\end{equation}

We next remove the past-cost terms. For $t'<t$, $c_{t'}$ is determined before $y_t$ is sampled, so:
\begin{equation}
\begin{aligned}
\mathbb E[c_{t'}\boldsymbol{g}_t] &= \mathbb E \Big[ c_{t'} \, \mathbb E_{y_t\sim\pi_{\boldsymbol{\theta}}(\cdot|\boldsymbol{s}_t)} [\boldsymbol{g}_t] \Big] \\
&= \mathbb E \Big[ c_{t'} \nabla_{\boldsymbol{\theta}} \sum_{y_t}\pi_{\boldsymbol{\theta}}(y_t|\boldsymbol{s}_t) \Big] = \mathbf{0}.
\end{aligned}
\end{equation}
Therefore, only current and future costs remain:
\begin{equation}
\nabla_{\boldsymbol{\theta}} J_{\mathrm{OPD}}(\boldsymbol{\theta}) = \mathbb E_{\boldsymbol{x},\boldsymbol{y}} \bigg[ \sum_{t=1}^T \Big( \sum_{t'=t}^T c_{t'} \Big) \boldsymbol{g}_t \bigg].
\end{equation}

Introducing the discounted return-to-go $G_t^\gamma := \sum_{t'=t}^T \gamma^{t'-t}c_{t'}$ for $\gamma\in[0,1]$, yields the practical estimator:
\begin{equation}
\nabla_{\boldsymbol{\theta}} J_{\mathrm{OPD}}^\gamma(\boldsymbol{\theta}) \approx \mathbb E_{\boldsymbol{x},\boldsymbol{y}} \bigg[ \sum_{t=1}^T G_t^\gamma \boldsymbol{g}_t \bigg].
\end{equation}
With $\gamma=0$, we have $G_t^0=c_t$, giving the token-local estimator:
\begin{equation}
\nabla_{\boldsymbol{\theta}} J_{\mathrm{OPD}}^{\gamma=0}(\boldsymbol{\theta}) \approx \mathbb E_{\boldsymbol{x},\boldsymbol{y}} \bigg[ \sum_{t=1}^T c_t \boldsymbol{g}_t \bigg].
\end{equation}
Equivalently, each sampled token receives the immediate local reward:
\begin{equation}
r_t = -c_t = \log\pi_{\boldsymbol{\phi}}(y_t|\boldsymbol{s}_t) - \log\pi_{\boldsymbol{\theta}}(y_t|\boldsymbol{s}_t),
\end{equation}
rather than a rollout-level return.

By truncating future costs, the $\gamma=0$ estimator serves as a biased approximation of the sequence-level KL divergence, making it theoretically more rigorous to interpret as a token-level RL process rather than strict distillation. This property, collapsing return to immediate cost, enabled the prefix rollouts to compute their updates on the fly without waiting for subsequent generations, and consequently eliminates the variance caused by the dependency on future positions.

\subsection{Local Surrogate Improvement and Descent Utility}
\label{app:descent-utility}

Here, we derive the optimization utility of the prefix in Sec~\ref{sec:grad_align}. In a local neighborhood around the parameter $\boldsymbol{\theta}\in\Theta$. Let $\boldsymbol{g} := \nabla_{\boldsymbol{\theta}} J_{\mathrm{OPD}}(\boldsymbol{\theta})$ denote the exact full-horizon gradient. We define $\boldsymbol{M}(\boldsymbol{\theta})$ as the local quadratic form representing the geometric curvature (e.g., the empirical Fisher Information Matrix):
\begin{equation}
\begin{aligned}
\boldsymbol{M}(\boldsymbol{\theta}) &\approx \mathbb{E}_{\boldsymbol{y}\sim\pi_{\boldsymbol{\theta}}} \Big[ \big(\nabla_{\boldsymbol{\theta}} \log \pi_{\boldsymbol{\theta}}(\boldsymbol{y})\big) \\
&\qquad\qquad \times \big(\nabla_{\boldsymbol{\theta}} \log \pi_{\boldsymbol{\theta}}(\boldsymbol{y})\big)^\top \Big].
\end{aligned}
\end{equation}
The second-order surrogate improvement of a parameter update step $\boldsymbol{\Delta}$ is given by its Taylor expansion up to the second order:
\begin{equation}
\mathcal{S}_{\boldsymbol{M}}(\boldsymbol{\Delta}) := \boldsymbol{g}^\top \boldsymbol{\Delta} - \frac{1}{2} \boldsymbol{\Delta}^\top \boldsymbol{M}(\boldsymbol{\theta}) \boldsymbol{\Delta}.
\end{equation}

Suppose we are restricted to updating the parameters along a specific surrogate direction $\boldsymbol{d} \neq \mathbf{0}$ with a step size $\eta \ge 0$. The update is $\boldsymbol{\Delta} = \eta \boldsymbol{d}$, and the surrogate improvement becomes a scalar quadratic function of $\eta$:
\begin{equation}
\mathcal{S}_{\boldsymbol{M}}(\eta \boldsymbol{d}) = \eta (\boldsymbol{g}^\top \boldsymbol{d}) - \frac{\eta^2}{2} (\boldsymbol{d}^\top \boldsymbol{M}(\boldsymbol{\theta}) \boldsymbol{d}).
\end{equation}
Taking the derivative with respect to $\eta$ and setting it to zero yields the unconstrained optimal step size. Since we require $\eta \ge 0$ to move strictly in the descent direction, the projected optimal step size is:
\begin{equation}
\eta^* = \frac{[\boldsymbol{g}^\top \boldsymbol{d}]_+}{\boldsymbol{d}^\top \boldsymbol{M}(\boldsymbol{\theta}) \boldsymbol{d}}.
\end{equation}
Substituting $\eta^*$ back into the surrogate objective gives the maximum utility achievable along direction $\boldsymbol{d}$:
\begin{equation}
\begin{aligned}
\mathcal{U}_{\boldsymbol{M}}(\boldsymbol{d}) &:= \max_{\eta \ge 0} \mathcal{S}_{\boldsymbol{M}}(\eta \boldsymbol{d}) \\
&= \frac{[\boldsymbol{g}^\top \boldsymbol{d}]_+ (\boldsymbol{g}^\top \boldsymbol{d})}{\boldsymbol{d}^\top \boldsymbol{M}(\boldsymbol{\theta}) \boldsymbol{d}} - \frac{[\boldsymbol{g}^\top \boldsymbol{d}]_+^2}{2 \, \boldsymbol{d}^\top \boldsymbol{M}(\boldsymbol{\theta}) \boldsymbol{d}} \\
&= \frac{[\boldsymbol{g}^\top \boldsymbol{d}]_+^2}{2 \, \boldsymbol{d}^\top \boldsymbol{M}(\boldsymbol{\theta}) \boldsymbol{d}}.
\end{aligned}
\end{equation}

If we instead do not restrict the update direction, we can find the globally optimal update $\boldsymbol{\Delta}^*$ by setting the gradient of $\mathcal{S}_{\boldsymbol{M}}(\boldsymbol{\Delta})$ to zero:
\begin{equation}
\begin{split}
\nabla_{\boldsymbol{\Delta}} \mathcal{S}_{\boldsymbol{M}}(\boldsymbol{\Delta}) &= \boldsymbol{g} - \boldsymbol{M}(\boldsymbol{\theta}) \boldsymbol{\Delta} = \mathbf{0} \\
&\implies \boldsymbol{\Delta}^* = \boldsymbol{M}(\boldsymbol{\theta})^{-1} \boldsymbol{g},
\end{split}
\end{equation}
which recovers exactly the natural gradient. Plugging $\boldsymbol{\Delta}^*$ into the improvement function yields the absolute maximum local utility:
\begin{equation}
\begin{aligned}
\mathcal{U}_{\boldsymbol{M}}^* &= \boldsymbol{g}^\top (\boldsymbol{M}(\boldsymbol{\theta})^{-1} \boldsymbol{g}) \\
&\quad - \frac{1}{2} (\boldsymbol{M}(\boldsymbol{\theta})^{-1} \boldsymbol{g})^\top \boldsymbol{M}(\boldsymbol{\theta}) (\boldsymbol{M}(\boldsymbol{\theta})^{-1} \boldsymbol{g}) \\
&= \frac{1}{2} \boldsymbol{g}^\top \boldsymbol{M}(\boldsymbol{\theta})^{-1} \boldsymbol{g}.
\end{aligned}
\end{equation}

To evaluate the quality of the surrogate direction $\boldsymbol{d}$, we calculate its relative utility ratio compared to the optimal natural gradient:
\begin{equation}
\mathcal{E}_{\boldsymbol{M}}(\boldsymbol{d}) := \frac{\mathcal{U}_{\boldsymbol{M}}(\boldsymbol{d})}{\mathcal{U}_{\boldsymbol{M}}^*} = \frac{[\boldsymbol{g}^\top \boldsymbol{d}]_+^2}{\big(\boldsymbol{g}^\top \boldsymbol{M}(\boldsymbol{\theta})^{-1} \boldsymbol{g}\big) \big(\boldsymbol{d}^\top \boldsymbol{M}(\boldsymbol{\theta}) \boldsymbol{d}\big)}.
\end{equation}
By defining the $\boldsymbol{M}$-weighted inner product $\langle \boldsymbol{u}, \boldsymbol{v} \rangle_{\boldsymbol{M}} = \boldsymbol{u}^\top \boldsymbol{M} \boldsymbol{v}$ and its induced norm $\|\boldsymbol{u}\|_{\boldsymbol{M}} = \sqrt{\boldsymbol{u}^\top \boldsymbol{M} \boldsymbol{u}}$, the cosine similarity between the candidate direction $\boldsymbol{d}$ and the optimal natural update $\boldsymbol{M}(\boldsymbol{\theta})^{-1} \boldsymbol{g}$ is:
\begin{equation}
\begin{aligned}
&\cos_{\boldsymbol{M}} \big(\boldsymbol{d}, \boldsymbol{M}(\boldsymbol{\theta})^{-1} \boldsymbol{g}\big) \\
=& \frac{\boldsymbol{d}^\top \boldsymbol{M}(\boldsymbol{\theta}) \big(\boldsymbol{M}(\boldsymbol{\theta})^{-1} \boldsymbol{g}\big)}{\|\boldsymbol{d}\|_{\boldsymbol{M}} \|\boldsymbol{M}(\boldsymbol{\theta})^{-1} \boldsymbol{g}\|_{\boldsymbol{M}}} \\
= &\frac{\boldsymbol{d}^\top \boldsymbol{g}}{\sqrt{\boldsymbol{d}^\top \boldsymbol{M}(\boldsymbol{\theta}) \boldsymbol{d}} \sqrt{\boldsymbol{g}^\top \boldsymbol{M}(\boldsymbol{\theta})^{-1} \boldsymbol{g}}}.
\end{aligned}
\end{equation}
Squaring this expression exactly recovers the relative utility ratio. Thus, for any descent direction ($\boldsymbol{g}^\top \boldsymbol{d} \ge 0$), we mathematically establish that:
\begin{equation}
\mathcal{E}_{\boldsymbol{M}}(\boldsymbol{d}) = \cos_{\boldsymbol{M}}^2 \big(\boldsymbol{d}, \boldsymbol{M}(\boldsymbol{\theta})^{-1} \boldsymbol{g}\big).
\end{equation}

The derivation above demonstrates that the relative descent utility of any proxy gradient is strictly governed by its cosine alignment with the natural gradient. Given a prefix gradient $\boldsymbol{d}_p$ and a full-rollout gradient $\boldsymbol{d}_f$, we can compare their informational value:
\begin{equation}
\frac{\mathcal{E}_{\boldsymbol{M}}(\boldsymbol{d}_p)}{\mathcal{E}_{\boldsymbol{M}}(\boldsymbol{d}_f)} = \Bigg( \frac{\cos_{\boldsymbol{M}}\big(\boldsymbol{d}_p, \boldsymbol{M}(\boldsymbol{\theta})^{-1} \boldsymbol{g}\big)}{\cos_{\boldsymbol{M}}\big(\boldsymbol{d}_f, \boldsymbol{M}(\boldsymbol{\theta})^{-1} \boldsymbol{g}\big)} \Bigg)^2.
\end{equation}
In the transparent case ($\boldsymbol{d}_f \approx \boldsymbol{M}(\boldsymbol{\theta})^{-1} \boldsymbol{g}$), the relative utility of truncating a rollout reduces to:
\begin{equation}
\mathcal{E}_{\boldsymbol{M}}(\boldsymbol{d}_p) \approx \cos_{\boldsymbol{M}}^2 (\boldsymbol{d}_p, \boldsymbol{d}_f).
\end{equation}
\section{Implementation Details}
\subsection{Training Configurations}
\label{app:train_details}
\begin{table}[t]
\centering
\setlength{\tabcolsep}{14pt} 
\renewcommand{\arraystretch}{1.2} 
\begin{tabular}{lc}
\toprule
\textbf{Hyper-parameter} & \textbf{Value} \\
\midrule
Train Batch Size& 128 \\
Micro Batch Size & 128 \\
Rollouts $n$ & 8 \\
Maximum Prompt Length & 2,048 \\
Maximum Response Length & 8,192 \\
Sampling Temperature & 1.0 \\
Top-$p$ & 1.0 \\
Learning Rate & 1e-6 \\
Optimization Steps & 500 \\
Clip Coefficient $\epsilon_{high}$ & 0.28 \\
\bottomrule
\end{tabular}
\caption{Implementation details for GRPO training. The same configuration is used for both math and code RL.}
\label{tab:grpo}
\end{table}
\begin{table}[t]
\centering
\setlength{\tabcolsep}{14pt} 
\renewcommand{\arraystretch}{1.2} 
\begin{tabular}{lc}
\toprule
\textbf{Hyper-parameter} & \textbf{Value} \\
\midrule
Train Batch Size& 1024 \\
Rollouts $n$ & 1 \\
Maximum Prompt Length & 2,048 \\
Maximum Response Length & 8,192 \\
Sampling Temperature & 1.0 \\
Top-$p$ & 1.0 \\
Learning Rate & 3e-6 \\
\bottomrule
\end{tabular}
\caption{Implementation details for OPD-family training}
\label{tab:opd}
\end{table}

\begin{table}[t]
\centering
\setlength{\tabcolsep}{14pt} 
\renewcommand{\arraystretch}{1.2} 
\begin{tabular}{lc}
\toprule
\textbf{Hyper-parameter} & \textbf{Value} \\
\midrule
Train Batch Size& 128 \\
Maximum Sequence Length & 10240 \\
Learning Rate & 1e-5 \\
Warm-up Ratio & 0.03 \\
LR scheduler & Cosine \\
\bottomrule
\end{tabular}
\caption{Implementation details for Offline KD family.}
\label{tab:sft}
\end{table}

We detailed our training hyperparameters in Table~\ref{tab:grpo}, Table~\ref{tab:opd} and Table~\ref{tab:sft}
For all experiments within the OPD family, we train the single-task setting for 30 optimization steps, whereas the multi-task and strong-to-weak settings are trained for 50 steps. For offline KD, these numbers are adjusted to 240 and 400 steps, respectively.

\subsection{Implementation details of Baselines}
\label{app:baseline_details}

Following~\citep{zhang2026fast}, we implement Fast OPD using the same linear window scheduling, which starts at 256 tokens and extends by an additional 256 tokens per step. For P-Align~\citep{liu2026long}, we strictly follow the original self-evaluation procedure and the binary-search-based sufficiency criterion. Since the original framework focused exclusively on mathematical reasoning, we adapt and extend the code block separation signs to operate at the logical line level. Furthermore, we observed that instruction-following capability becomes a bottleneck when prompting the student to continue generation using the teacher's prefix as a user-side input draft, which consistently degrades performance. To mitigate this and ensure data quality, we instead prefill the teacher-generated segment as a direct prefix during student generation. We directly adopted standard setups for all other baselines.

\section{Additional Analysis Results}

\label{app:analysis}
\subsection{Loss Distribution on Prefix}
\begin{figure*}[t]
  \includegraphics[width=0.49\linewidth]{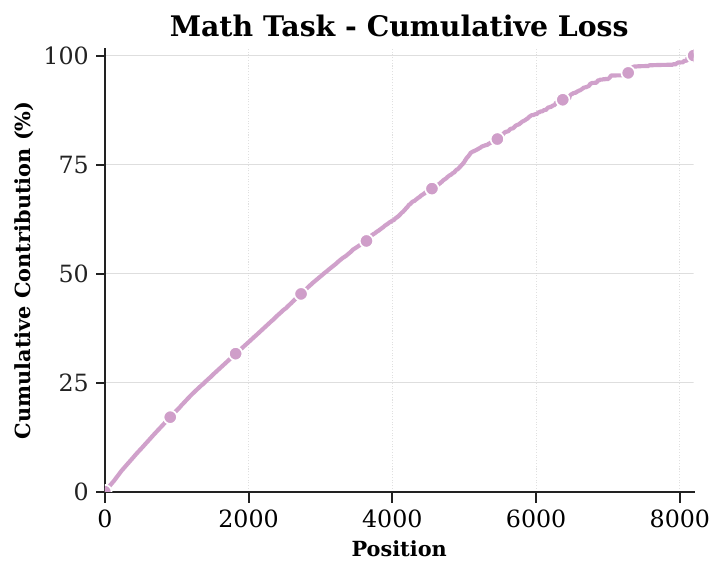} \hfill
  \includegraphics[width=0.49\linewidth]{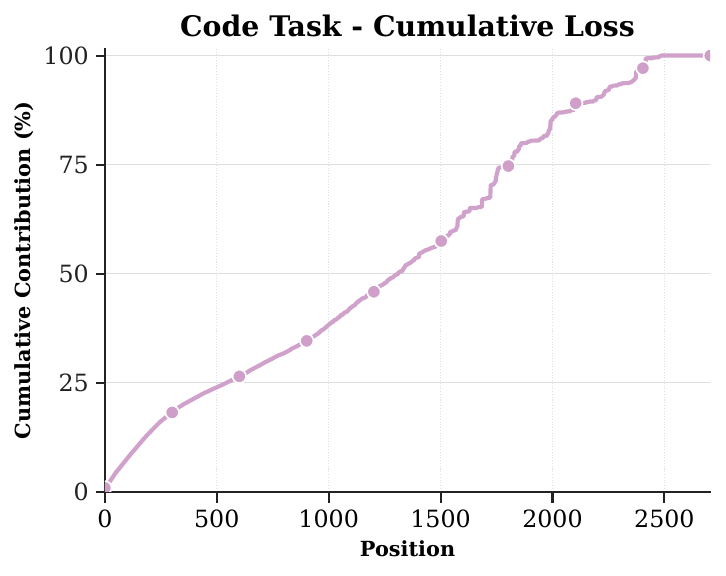}
\caption{
Cumulative distribution of token-level OPD loss over response positions on Qwen3-1.7B setting. 
The x-axis denotes the token position, and the y-axis reports the cumulative percentage of total loss accumulated up to that position. 
Results are shown separately for math and code tasks, illustrating loss distributed along student-generated rollouts are rather uniform than concentrated.
}
\label{fig:loss_distribution}
\end{figure*}

We consider the token level loss additionally as for reference, illustrated in Figure~\ref{fig:loss_distribution}, where we observed a rather uniform distribution as in cumulative way. 

\subsection{The Cascading Effect on Suffix Loss}
\begin{figure*}[t]
  \includegraphics[height=0.35\linewidth, width=0.49\linewidth]{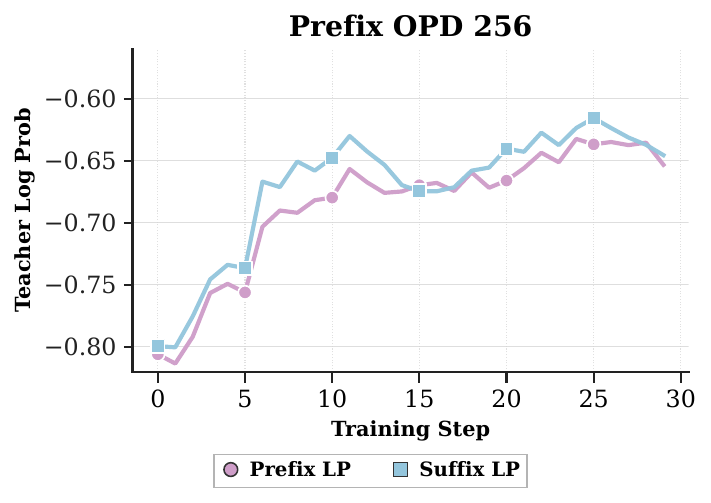} \hfill
  \includegraphics[height=0.35\linewidth, width=0.49\linewidth]{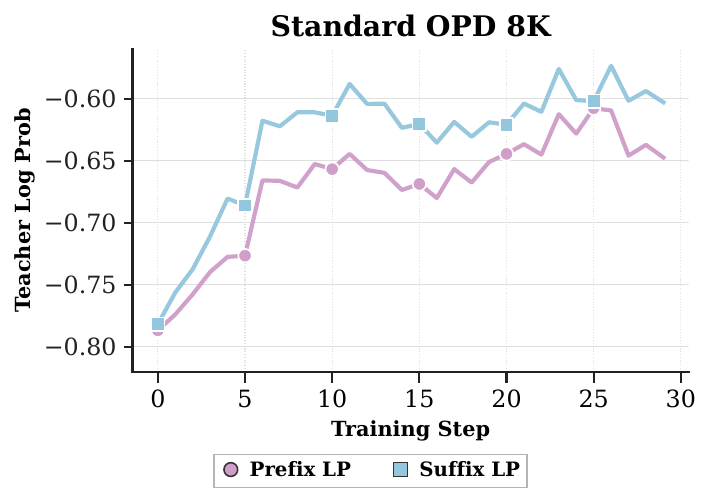}
\caption{
Prefix and suffix teacher log probability dynamics on Qwen3-1.7B setting. 
We compare Prefix OPD with a 256-token training window and standard full-rollout OPD with an 8K-token horizon. 
For each training step, teacher log probability is measured separately on the prefix and suffix regions of the generated response, showing that prefix optimization can also improve the downstream suffix region with a considerable extent.
}
\label{fig:cascading_effect}
\end{figure*}

We further conducted an analysis experiment, where we generate the full rollout, but applied a loss mask on its prefix and monitor the training process. Our empirical findings in Figure~\ref{fig:cascading_effect}reveal a compelling cascading effect: optimizing the prefix implicitly yet effectively drives down the suffix loss. While this confirms the prefix's outsized influence on the trajectory, computing the full sequence merely to mask it defeats the computational purpose of truncation. This underscores the necessity of our proposed approach: an online, training-dynamic-aware mechanism that can adaptively identify the effective prefix boundary in practice, without relying on expensive full-rollout references."

\section{Additional Results}

\begin{table*}[!t]
    \centering
    \small
    \renewcommand{\arraystretch}{1.3}
    \setlength{\tabcolsep}{5pt}
    \begin{tabular}{lcccccccc}
    \toprule
    \multirow{3}{*}{\textbf{Method}}
    & \multicolumn{3}{c}{\textbf{Mathematics}}
    & \multicolumn{3}{c}{\textbf{Code}}
    & \multirow{3}{*}{\textbf{Avg.}}
    & \multirow{3}{*}{\textbf{EFLOPs} $\downarrow$} \\
    \cmidrule(lr){2-4}\cmidrule(lr){5-7}
    & AIME 24 & AIME 25 & Beyond AIME & HumanEval+ & MBPP+ & LCB v6 & & \\
    & \texttt{\footnotesize avg@16} & \texttt{\footnotesize avg@16} & \texttt{\footnotesize avg@8}
    & \texttt{\footnotesize avg@4}  & \texttt{\footnotesize avg@4}  & \texttt{\footnotesize avg@4} & & \\
    \midrule
    Initial Model & 13.3 & \phantom{0}7.9 & \phantom{0}4.5 & 57.1 & 40.2 & 13.1 & 22.7 & -- \\
    Teacher       & 76.0 & 61.9           & 44.8           & 92.6 & 79.4 & 40.9 & 65.9 & -- \\
    \midrule
     ADWIN $\rho=0.5$  & 32.3 & 24.8 & \textbf{13.9} & 23.7 & 68.7 & 29.1 & 39.1 & \textbf{0.6} \\
     ADWIN $\rho=0.6$   & 32.5 & 25.4 & \textbf{13.9} & 23.9 & 69.3 & 27.4 & 39.1 & 0.8 \\
     ADWIN $\rho=\rho^*$   & \textbf{34.2} & 26.9 & 13.4 & \textbf{24.8} & \textbf{72.4} & 27.7 & \textbf{40.3} & 1.2 \\
     ADWIN $\rho=0.8$   & 32.1 & \textbf{29.8} & 11.4 & 24.4 & 70.6 & \textbf{29.1} & \textbf{40.3} & 1.5 \\
    \bottomrule
    \end{tabular}
    \caption{Sensitivity analysis of the admissibility threshold $\rho$ in ADWIN across the strong-to-weak setting. Lower thresholds reduce compute but may accept insufficiently aligned prefixes, while stricter thresholds increase the training horizon without yielding consistent accuracy gains. 
The default threshold $\rho^*$ lies near this transition region, attaining the highest average accuracy among the evaluated choices (less compute than 0.8).}   
    \label{tab:ablate_rho}
\end{table*}

\begin{table*}[!t]
    \centering
    \small
    \renewcommand{\arraystretch}{1.1}
    \setlength{\tabcolsep}{5pt}
    \begin{tabular}{lcccccccc}
    \toprule
    \multirow{3}{*}{\textbf{Method}}
    & \multicolumn{3}{c}{\textbf{Mathematics}}
    & \multicolumn{3}{c}{\textbf{Code}}
    & \multirow{3}{*}{\textbf{Avg.}}
    & \multirow{3}{*}{\textbf{EFLOPs} $\downarrow$} \\
    \cmidrule(lr){2-4}\cmidrule(lr){5-7}
    & AIME 24 & AIME 25 & Beyond AIME & HumanEval+ & MBPP+ & LCB v6 & & \\
    & \texttt{\footnotesize avg@16} & \texttt{\footnotesize avg@16} & \texttt{\footnotesize avg@8}
    & \texttt{\footnotesize avg@4}  & \texttt{\footnotesize avg@4}  & \texttt{\footnotesize avg@4} & & \\
    \midrule
    Prefix OPD-128  & 22.5 & 15.2 & \phantom{0}8.0 & 68.3 & 61.0 & 21.5 & 32.8 & \textbf{0.4} \\
    Prefix OPD-256  & 31.0 & 22.1 & 10.8 & 67.6 & 59.6 & 21.3 & 35.4 & 0.5 \\
    Prefix OPD-512  & 31.7 & 26.0 & \textbf{13.9} & \textbf{75.2} & 65.6 & 23.3 & 39.3 & 0.7 \\
    Prefix OPD-1024 & 34.1 & 24.0 & 13.8 & 74.8 & 66.8 & 26.9 & 40.1 & 1.0 \\
    Prefix OPD-2048 & \textbf{34.4} & 24.8 & 12.7 & 72.2 & 64.0 & 21.4 & 38.3 & 1.5 \\
    \midrule
    ADWIN ($\rho^*$) & 34.2 & \textbf{26.9} & 13.4 & 24.8 & \textbf{72.4} & \textbf{27.7} & \textbf{40.3} & 1.2 \\
    \bottomrule
    \end{tabular}
    \caption{Detailed prefix ablation results in ADWIN across strong-to-weak setting.}   
    \label{tab:detailed_prefix_ablate}
\end{table*}

We use $\rho^*=\sqrt{2}/2$ as the default admissibility threshold in ADWIN, and provide additional sensitivity results to show how varying this threshold affects the accuracy--efficiency trade-off.

\section{Instruction Templates}
\label{app:prompt}

The instructions we used for both training and evaluation are detailed below.

\vspace{0.5em}
\noindent\textbf{Prompt Template for Math Reasoning:}
\nopagebreak

\noindent
{\ttfamily
    \textbf{\underline{User Message}}\\
    <|im\_start|>user\\
    \{question\}\\
    Please reason step by step, and put your final answer within \textbackslash{}boxed\{\}.<|im\_end|>\\
    <|im\_start|>assistant\\
}
\nopagebreak
{\ttfamily
    \textbf{\underline{Assistant Message}}\\
    \{response\}\\
}

\vspace{1.5em} 

\noindent\textbf{Prompt Template for Code Generation:}
\nopagebreak

\noindent
{\ttfamily
    \textbf{\underline{User Message}}\\
    <|im\_start|>user\\
    \{question\}\\
    Write Python code to solve the problem. Present the code in\\
    \verb|```|python\\
    Your code\\
    \verb|```|\\
    at the end.\\
    You need to think first then write the Python code.<|im\_end|>\\
    <|im\_start|>assistant\\
}
\nopagebreak
{\ttfamily
    \textbf{\underline{Assistant Message}}\\
    \{response\}\\
}
\section{Qualitative Cases of Student Drift}
\label{app:case}
We provide qualitative cases here where student context drift is severe, and the teacher's supervision provides noise rather than reliable guidance. Green denotes a positive teacher feedback on that token while red denotes penalize, with the color intensity reflecting the magnitude of the feedback value.

\begin{figure*}[t]
    \centering
    \includegraphics[width=1.0\linewidth]{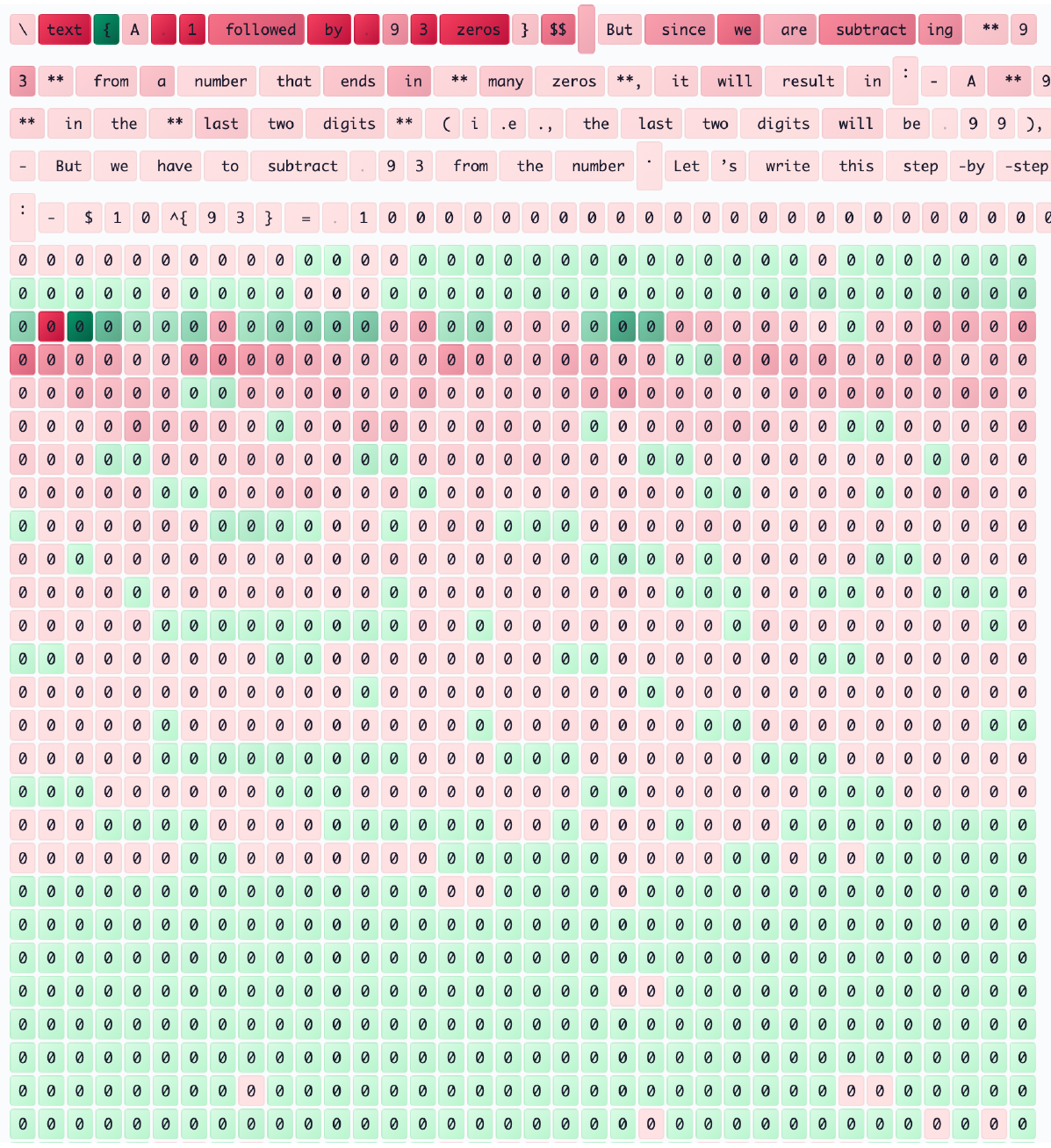}
    \caption{
Qualitative case of student context drift into a repetitive loop on mathematical reasoning. The student model enters a meaningless repetition sequence, where teacher distribution allocates a significant proportion of positive reward signals (denoted in green) over these degenerate tokens.
}
    \label{fig:adwin_framework}
\end{figure*}

\begin{figure*}[t]
    \centering
    \includegraphics[width=1.0\linewidth]{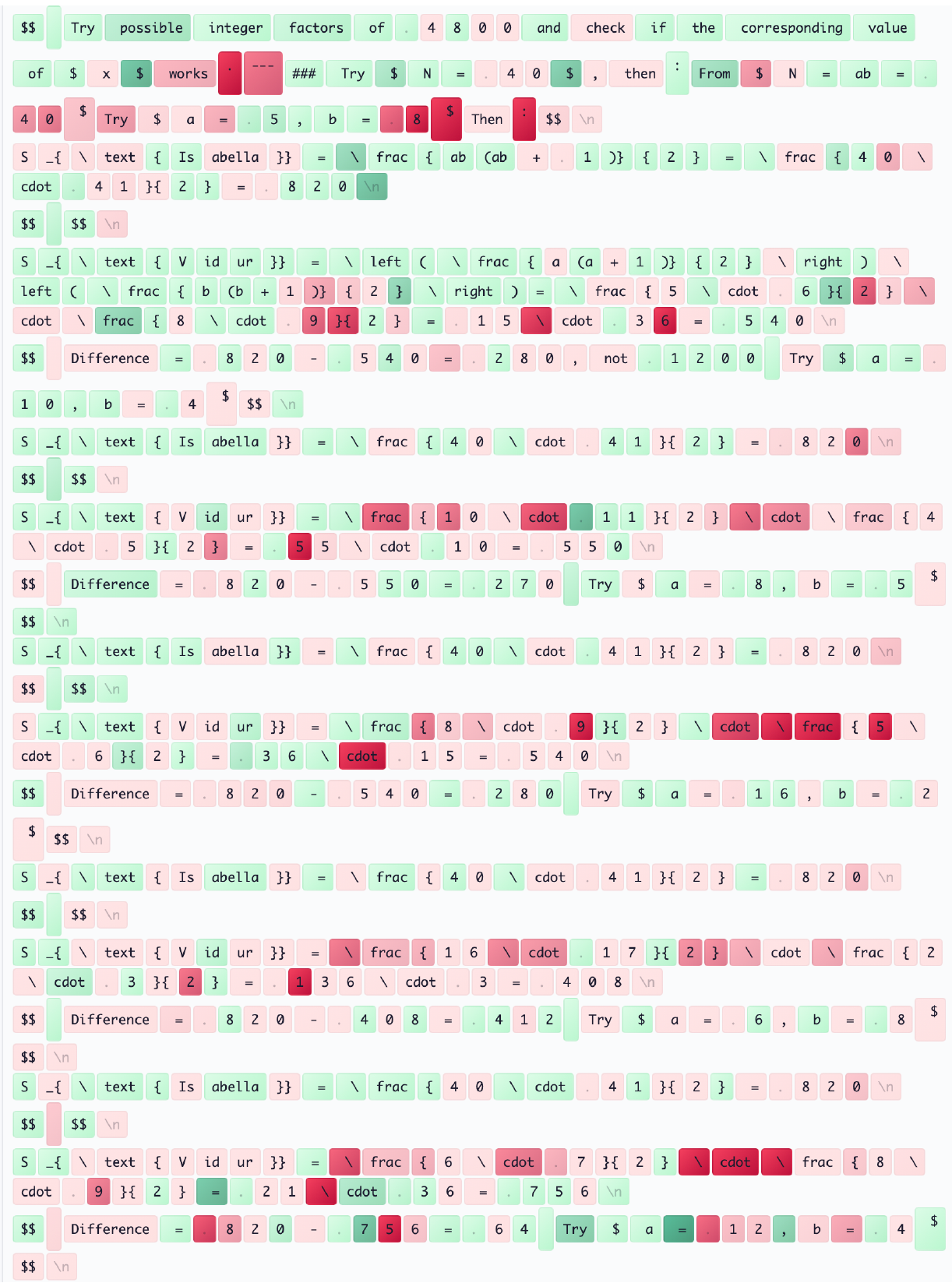} 
    \caption{
    The student model enters a trial-and-error loop of evaluating different candidate integer factors ($a$ and $b$), where the teacher provides a non-trivial proportion positive feedback (denoted in green).
    }
    \label{fig:adwin_framework}
\end{figure*}

\end{document}